\documentclass[12pt]{report}
\usepackage[T1]{fontenc}
\usepackage[utf8]{inputenc}
\usepackage[a4paper, margin=1in]{geometry}
\usepackage{tabto}
\usepackage{float}
\usepackage{color}   %May be necessary if you want to color links
\usepackage{hyperref}
\usepackage[toc,page,title,titletoc]{appendix}
\usepackage[
    backend=biber,
    style=apa,
    sorting=nyt
]{biblatex}
\usepackage[parfill]{parskip}
\hypersetup{
    colorlinks=true, %set true if you want colored links
    linktoc=all,     %set to all if you want both sections and subsections linked
    linkcolor=blue,  %choose some color if you want links to stand out
    citecolor=black,
    filecolor=black,
    linkcolor=black,
    urlcolor=black
}
\usepackage{multirow}
\usepackage{makecell}
\usepackage{array}
\usepackage{tabularx}
\usepackage{siunitx} % For aligning numbers by decimal point
\usepackage{booktabs} % For better-looking horizontal separators
\usepackage{graphicx}
\usepackage{amsmath}
\usepackage{amssymb}
\usepackage{caption}
\usepackage[table,xcdraw]{xcolor}
\graphicspath{ {images/} }
\usepackage{mdframed}
\definecolor{bg}{rgb}{0.95,0.95,0.95}
\usepackage{enumitem}

\usepackage{times}
\usepackage{setspace} % Include the setspace package
\usepackage{listings}

\usepackage{algorithm}
\usepackage{algpseudocode}

\usepackage{tikz}
\usetikzlibrary{tikzmark, arrows.meta}

\renewenvironment{abstract}
 {\small
  \begin{center}
  \bfseries \abstractname\vspace{-.5em}\vspace{0pt}
  \end{center}
  \addcontentsline{toc}{chapter}{\abstractname} % Add to ToC
  \list{}{%
    \setlength{\leftmargin}{0pt}
    \setlength{\rightmargin}{\leftmargin}%
  }%
  \item\relax}
 {\endlist}

\newenvironment{acknowledgements}
 {\small
  \begin{center}
  \bfseries Acknowledgements\vspace{-.5em}\vspace{0pt}
  \end{center}
  \addcontentsline{toc}{chapter}{Acknowledgements} % Add to ToC
  \list{}{%
    \setlength{\leftmargin}{0pt}
    \setlength{\rightmargin}{\leftmargin}%
  }%
  \item\relax}
 {\endlist}

 \renewcommand\tab[1][10mm]{\hspace*{#1}}

\newcommand{\projecttype}{B.Comp. Dissertation}
\newcommand{\authorname}{Yeoh Zhong Han Ervin}
\newcommand{\projecttitle}{From Skeletons to Pixels: Few-Shot Precise Event Spotting via Representation and Prediction Distillation}
\newcommand{\academicyear}{2025/2026}

\begin{document}

\newgeometry{left=3cm,right=3cm,top=3cm,bottom=3cm}
\begin{titlepage}
    \centering
    {\large \projecttype}
    \vfill
    {\LARGE\bfseries \projecttitle\par}
    \vspace{2cm}
    {\large By\par \authorname}
    \vfill
    Department of Business Analytics\par
    School of Computing\par
    National University of Singapore\par
    \vspace{1cm}
% Bottom of the page
    {\normalsize \academicyear\par}
\end{titlepage}
\restoregeometry
\pagebreak

\newgeometry{left=3cm,right=3cm,top=3cm,bottom=3cm}
\begin{titlepage}
    {\centering
    {\large \projecttype}
    \vfill
    {\LARGE\bfseries \projecttitle\par}
    \vspace{2cm}
    {\large By\par \authorname}
    \vfill
    {\normalsize Department of Computer Science\par
    School of Computing\par
    National University of Singapore\par}
    \vspace{1cm}
% Bottom of the page
    {\normalsize \academicyear\par}
    }
    \vspace{1cm}
    Project ID: H403120 \par
    Supervisor(s): Michael Jiang Kan \par
    \vspace{1cm}
    Deliverables: \par
    \begin{list}{\quad}{}
        \item Report: 1 Volume
        \item Manual: 1 Volume
        \item Program: 1 Diskette 
        \item Database: 1 Diskette
    \end{list}
\end{titlepage}
\restoregeometry
\pagebreak

\pagenumbering{roman}
\begin{abstract}
\noindent % Prevents indenting at the start of the paragraph
Precise Event Spotting (PES) is essential in fast-paced sports such as tennis, where fine-grained events occur within very short temporal windows. Accurate frame-level localization is challenging due to motion blur, subtle action differences, and limited annotated data. We study two complementary distillation strategies for few-shot PES: 1) \textbf{Adaptive Weight Distillation (AWD)}, a prediction-level method that adaptively weights teacher supervision on unlabeled data, and 2) \textbf{Annealed Multimodal Distillation for Few-Shot Event Detection (AMD-FED)}, a representation-level framework that transfers robust skeleton knowledge into visual modalities through annealed pseudo-labeling. Both methods leverage multimodal distillation to improve generalization under limited supervision. We evaluate both methods on $F^3$Set-Tennis(sub) under few-shot $k$-clip settings, where they consistently outperform single-modality baselines and previous PES approaches. After observing the superior performance of representation-level distillation on tennis, we further validate AMD-FED on a second sports dataset, Figure Skating, where it also demonstrates robust performance in the $k$-clip scenario. These results highlight the effectiveness of multimodal distillation, and in particular representation-level transfer, for few-shot precise event spotting.

\vspace{5mm}
Subject Descriptors:
\begin{list}{\quad}{}
    \item I.2.10 Vision and Scene Understanding
    \item I.4 Image Processing and Computer Vision
\end{list}

Keywords: \\
\tab Precise Event Spotting, few-shot learning, multimodal knowledge distillation, semi-supervised learning, action recognition
\end{abstract}
\pagebreak

\begin{acknowledgements}
\noindent % Prevents indenting at the start of the paragraph
I would like to sincerely thank Michael and Zhaoyu for guiding me through both UROP and FYP. Their support and mentorship made this journey possible. Looking back, going from barely understanding what a neural network was to being able to complete this project independently feels incredible, and I could not have done it without them.

I also want to thank my friends who helped me along the way, in particular Charlton Teo, for helping me througout the entire UROP journey and continuing to inspire my research journey.
\end{acknowledgements}
\pagebreak

% only show chapters and sections in the ToC
\setcounter{tocdepth}{1}
\tableofcontents
\clearpage

\pagenumbering{arabic}
\setcounter{page}{1}
\chapter{Introduction}
\label{chap:Introduction}

Precise Event Spotting (PES) plays a critical role in modern sports analytics by enabling the accurate temporal localization of key events within continuous video streams. In professional sports, PES supports applications such as automated highlight generation \autocite{javed2016efficient}, fine-grained performance analysis for athletes and coaches \autocite{chen2024interpretable,liu2023insight}, and real-time decision support for refereeing systems \autocite{ma2024refereeing}. In fast-paced rally sports such as tennis, where events such as hits, serves, and bounces occur within milliseconds, precise temporal accuracy is essential for downstream tasks including tactic analysis \autocite{decroos2018automatic}, player workload estimation \autocite{antonini2024engineering,black2016monitoring}, and broadcast augmentation. Consequently, PES serves as a fundamental component for scalable, data-driven sports understanding systems.

\begin{figure}[htbp]
	\centering
	% Put your figure at images/pes-data.png
	\includegraphics[width=0.9\linewidth]{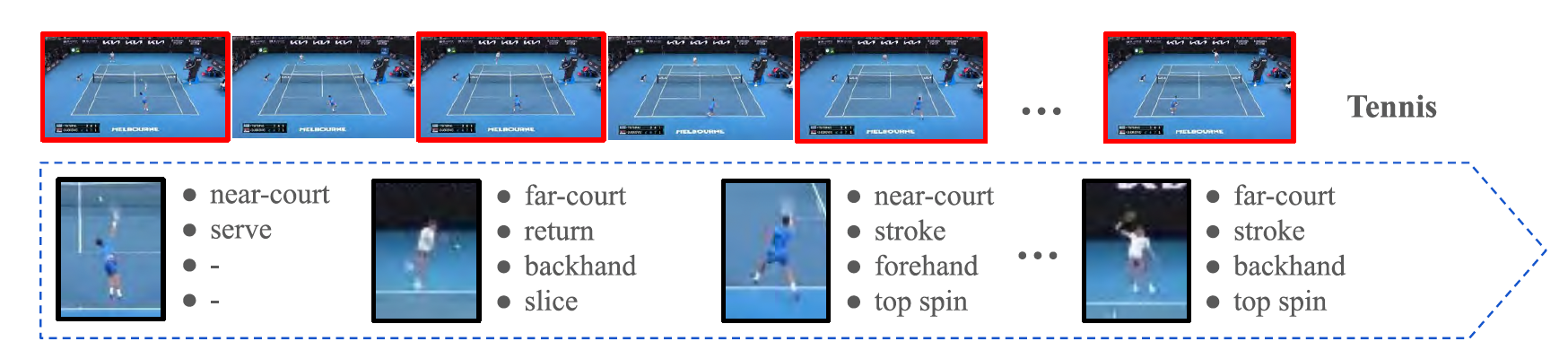}
	\caption{Example of fine-grained event annotation in fast-paced sports, with event timestamps highlighted in red, adopted from \cite{liu2025f3set}.}
	\label{fig:pes-data}
\end{figure}

Large-scale datasets \autocite{bian2024p2anet,liu2025f3set,shao2020finegym,wang2023shuttleset,xu2022finediving} have been introduced to advance research in Precise Event Spotting (PES), but they typically require dense temporal annotations (see Figure~\ref{fig:pes-data}), incurring substantial human effort and time costs. Moreover, fully supervised training paradigms often suffer from limited generalization to unseen scenarios, long training durations, and a tendency to overfit when annotations are scarce. \textbf{In this work, we focus on a few-shot setting, where the model must detect events and classify actions given only $k$ video clips in total, simulating realistic low-data conditions.}

Existing approaches \autocite{hong2022spotting,liu2025f3set} also predominantly rely on a single modality, further exacerbating overfitting in few-shot settings. From a modeling perspective, fine-grained action detection methods can be broadly categorized into end-to-end visual models \autocite{carreira2017quo,feichtenhofer2019slowfast,lin2019tsm,wang2018temporal} and skeleton-based approaches \autocite{liu2020disentangling}. Visual models operate directly on raw RGB frames, often incorporating optical flow to improve robustness, but their heavy dependence on pixel-level patterns causes them to latch onto static visual cues rather than learning generalizable motion representations, particularly in low-data regimes. Skeleton-based methods, by contrast, leverage structured pose representations to model human motion and exhibit stronger generalization, yet they discard rich visual context and rely on accurate pose estimation. In fast-paced sports scenarios, even state-of-the-art pose detectors \autocite{sun2019deep,wu2019detectron2} frequently fail due to motion blur, occlusions, and rapid movements, often in frames critical for event recognition. As a result, both paradigms struggle to generalize to fine-grained sports actions under limited supervision.

While knowledge distillation has been widely explored to improve few-shot learning, prior work predominantly focuses on multi-class classification tasks on datasets such as CIFAR \autocite{krizhevsky2009learning} and MNIST \autocite{lecun2010mnist}, or on coarse-grained action recognition videos \autocite{bock2023wear,damen2018epic,grauman2022ego4d,chen2026action100m}, where either representation-level distillation \autocite{hatano2024multimodal,du2023semisupervisedlearningweightawaredistillation} or prediction-layer distillation \autocite{kontonis2023slamstudentlabelmixingdistillation,iliopoulos2022weighted} is applied. In contrast, the challenges of Precise Event Spotting, including short temporal windows, multi-label per-frame annotations, and subtle motion cues, have rarely been addressed within a distillation framework.

In this work, we embed both representation-level and prediction-layer distillation strategies into the PES setting, enabling robust few-shot learning by transferring knowledge from skeleton-based motion priors to visual modalities. Unlike prior distillation work that mainly targets multi-class classification or coarse-grained action recognition, our approach explicitly addresses the challenges of Precise Event Spotting, including short temporal windows, multi-label per-frame annotations, and subtle motion cues. We also introduce improvements specific to each distillation paradigm to enhance performance in this domain.

To this end, we propose two complementary frameworks:

\begin{itemize}
	\item \textbf{Annealed Multimodal Distillation for Few-Shot Event Detection (AMD-FED)} implements representation-level distillation. We improve the robustness of the skeleton teacher by introducing an \textbf{Annealed Pseudo-Labeling Strategy} during semi-supervised pretraining, where unlabeled samples are gradually incorporated using a time-dependent weighting schedule. This mitigates noise from early incorrect predictions and yields a more reliable teacher for downstream distillation to RGB and optical-flow students.

	\item \textbf{Adaptive Weight Distillation (AWD)} extends prediction-layer distillation to the multi-label PES setting. We design a tennis-specific confidence estimation and post-processing procedure, which adapts the teacher's supervision based on per-frame and per-clip reliability, enabling effective knowledge transfer even in challenging multi-label scenarios.
\end{itemize}

We evaluate both methods on $F^3$Set-Tennis(sub) under few-shot settings. On this benchmark, AWD and AMD-FED consistently outperform single-modality baselines and prior PES methods, while AMD-FED achieves the best overall results with \textbf{80.24\%} and \textbf{83.84\%} Edit score in the 50-clip and 100-clip settings, exceeding the strongest single-modality baseline by 9.17 and 7.19 percentage points, respectively. Motivated by the superior performance of representation-level distillation on tennis, we further evaluate AMD-FED on the Figure Skating dataset and observe that it consistently outperforms the strongest competing baseline under both few-shot settings. These results demonstrate the effectiveness of distillation for few-shot PES, and in particular the advantage and robustness of representation-level transfer for fine-grained event spotting.

\chapter{Related Work}
\label{chap:RelatedWork}

\textbf{Few-Shot Temporal Action Detection} Few-shot learning (FSL) aims to enable models to generalize to novel classes using only a limited number of labeled examples, addressing the scalability challenges of fully supervised learning \autocite{fei2006one,salakhutdinov2013learning}. A predominant paradigm for FSL is supervised meta-learning, where models are trained across episodic tasks to rapidly adapt to unseen classes \autocite{hinton1987fast,finn2017maml,thrun1998learning}. Existing meta-learning approaches can be broadly categorized into three groups: metric-based methods, which learn embedding spaces that facilitate similarity-based classification \autocite{snell2017prototypical,sung2018learning,vinyals2016matching}; memory-based methods, which store and retrieve task-specific knowledge to support generalization \autocite{santoro2016meta,oreshkin2018tadam,mishra2018snail,munkhdalai2017metanetworks}; and gradient-based methods, which learn optimization strategies that enable fast adaptation of a base learner to new tasks \autocite{finn2017maml,ravi2017optimization,lee2018gradient,grant2018recasting,zhang2018metagan}.

Extending FSL to temporal action detection (TAD) introduces additional challenges due to the need for precise temporal localization. Early work integrated sliding-window proposals with metric learning frameworks to perform few-shot TAD in untrimmed videos \autocite{vinyals2016matching,yang2018one}. More recently, Transformer-based architectures have been adopted to better align support and query features in untrimmed video settings \autocite{nag2021few}. Unlike traditional $k$-shot formulations, where each support example contains a single action instance and background \autocite{nag2021few}, several works, including ours, evaluate few-shot temporal event detection using sparsely annotated clips that may contain multiple events and timestamps \autocite{hong2021video}. Despite these advances, existing FSL-based action detection methods often struggle in fine-grained sports scenarios, where events are brief, visually subtle, and subject to high intra-class variation. In this work, we address few-shot fine-grained sports event detection by emphasizing precise temporal localization and leveraging multimodal distillation to improve robustness and adaptation under limited labeled data.

\textbf{Knowledge Distillation} Knowledge distillation (KD) serves as a key approach for transferring information from a large, well-trained teacher network to a smaller or less capable student network. Traditional KD methods \autocite{hinton2015distilling} operate under a fully supervised setting, where distillation is performed on the labeled training data of the teacher model, rather than on new, unlabeled data. Extensions such as the variational framework \autocite{ahn2019variational} aim to further align the student's outputs and intermediate representations \autocite{chen2020learning} with those of the teacher.

In cases where labeled data is scarce and unlabeled samples dominate, the reliability of teacher-generated pseudo-labels becomes uncertain. Studies \autocite{stanton2021knowledge} have shown that over-reliance on noisy pseudo-labels can degrade student performance. To mitigate this, recent works \autocite{dehghani2017fidelity,lang2022training,iliopoulos2022weighted} propose confidence-based filtering or reweighting of pseudo-labels using uncertainty estimates, such as entropy or margin scores, or explicitly model the teacher's noise to adjust the student's learning objective. These approaches highlight the importance of robust distillation techniques for scenarios involving unlabeled or weakly labeled data. In this work, we evaluate the performance of pseudo-label-based approaches in comparison to representation-level distillation.

\chapter{Our Approach}
\label{chap:Approach}

We study the problem of few-shot precise event detection and propose two distillation-based frameworks that adapt knowledge transfer techniques, commonly used in action recognition, to the Precise Event Spotting (PES) setting.

Specifically, we consider two standard distillation paradigms: \emph{(i) representation-level distillation} and \emph{(ii) prediction-layer distillation}. While both paradigms have been extensively explored for video-level action recognition, their application to fine-grained, timestamp-level event detection remains underexplored.

Based on these paradigms, we introduce:
\begin{itemize}
    \item \textbf{AMD-FED}, which performs \emph{representation-level multimodal distillation}, transferring domain-adapted motion features from skeletons to visual modalities; and
    \item \textbf{AWD}, which performs \emph{prediction-layer distillation} by directly supervising student predictions using weighted teacher outputs.
\end{itemize}

An overview of the AMD-FED framework is illustrated in Figure~\ref{fig:framework}. Unless otherwise stated, AMD-FED serves as our primary model, while AWD is used as a complementary distillation strategy for comparison.
For clarity of presentation, we use tennis as the running case study throughout this chapter, particularly when describing label structure and post-processing logic; however, the overall framework is intended to generalize to other fine-grained sports datasets.

\begin{figure*}[t]
    \centering
    \includegraphics[width=\textwidth]{framework.png}
    \caption{The framework of our proposed method. AMD-FED has three training stages: 1. learning domain-specific skeleton features from both labeled and unlabeled data, 2. distilling the skeleton feature (frozen) into student RGB and optical flow encoders through self-supervised training using both labeled and unlabeled data, and 3. few-shot learning for adapting to the temporal event detection task in target sports using only the labeled data.}
    \label{fig:framework}
\end{figure*}

\section{Problem Statement}
\label{sec:ProblemStatement}

We study the few-shot Precise Event Spotting (PES) task under limited supervision, following the standard formulation adopted in prior work \autocite{liu2026umegnet}. Given an input video clip $\mathbf{X} \in \mathbb{R}^{T \times H \times W \times 3}$ consisting of $T$ RGB frames with spatial resolution $H \times W$, the objective is to localize a sequence of $M$ events in time, represented as $\{(E_1, t_1), \ldots, (E_M, t_M)\}$. Here, $E_i$ denotes the event category drawn from $C$ possible classes, and $t_i$ corresponds to the temporal timestamp of the $i$-th event.

We consider a target dataset $\mathcal{D}$ comprising $|\mathcal{D}|$ video clips, partitioned into a labeled subset $\mathcal{D}_{\text{label}} \subset \mathcal{D}$ and an unlabeled subset $\mathcal{D}_{\text{unlabel}} \subset \mathcal{D}$, such that $\mathcal{D} = \mathcal{D}_{\text{label}} \cup \mathcal{D}_{\text{unlabel}}$.

We specifically focus on the few-shot PES setting, where only a small number of labeled video clips, $k = |\mathcal{D}_{\text{label}}|$, are available for training. We refer to this regime as the \emph{$k$-clip setting}.

\section{AMD-FED: Representation-Level Multimodal Distillation}
\label{sec:AMDFED}

\subsection{Domain-Specific Feature Pretraining with Unlabeled Data}
\label{sec:DomainSpecificFeaturePretraining}

In Stage~I, we focus on learning \emph{domain-adapted and discriminative motion representations} by explicitly incorporating unlabeled data, which constitutes the main novelty of our framework. In contrast to existing few-shot event spotting approaches that rely solely on limited labeled samples, we introduce a semi-supervised pretraining strategy that leverages abundant unlabeled skeleton sequences available in the target domain.

Given an input pose sequence $X_{\text{pose}} \in \mathbb{R}^{T \times P \times V \times 2}$ (extracted by ~\autocite{sun2019deep}), where $T$ denotes the number of frames, $P$ the number of persons, and $V$ the number of joints with $(x,y)$ coordinates, we employ a skeleton-based teacher encoder $\varepsilon_{\text{pose}}^{\text{teacher}}$ consisting of a GCN backbone followed by a bidirectional GRU. This design enables the model to capture both spatial joint correlations and temporal motion dynamics. The resulting pose representation is
\begin{equation}
F_{\text{pose}} = \varepsilon_{\text{pose}}^{\text{teacher}}(X_{\text{pose}}).
\end{equation}

The pose features are fed into a temporal event detector with coarse-grained and fine-grained prediction branches to produce dense event predictions at each timestamp.

\noindent\textbf{Semi-supervised training with pseudo-labeling.}
Training is conducted in two phases to ensure stability. During the first 30 epochs, the model is optimized using labeled data only. We then select the \emph{best-performing checkpoint} from this warm-up phase and use it to initialize the subsequent semi-supervised training. Afterward, labeled and unlabeled samples are mixed with a 1:1 sampling ratio. Pseudo-labels for unlabeled samples are generated online using the model's own predictions and are used as supervisory signals.

This progressive introduction of unlabeled data, together with an epoch-dependent weighting schedule, alleviates the impact of noisy pseudo-labels while effectively expanding the training set. As a result, the model learns more robust domain-specific motion representations, which are crucial for few-shot event spotting.

\noindent\textbf{Labeled Coarse-grained loss.}
The coarse-grained branch predicts whether any event occurs at each timestamp. For labeled samples, let $\hat{p}_{b,t,c}$ denote the predicted probability for coarse class $c \in \{0,1\}$ at batch index $b$ and timestamp $t$, and let $y_{b,t,c}$ be the corresponding one-hot ground-truth label. The coarse-grained loss for labeled data is computed using standard cross-entropy:
\begin{equation}
L_{\text{coarse}}^{\text{lab}}
= - \frac{1}{B \times T}
\sum_{b=1}^{B} \sum_{t=1}^{T}
\log \hat{p}_{b,t,c}.
\end{equation}

\noindent\textbf{Labeled Fine-grained loss.}
The fine-grained branch predicts multiple event categories at each timestamp. Let $\hat{Y}_{b,t,c}$ denote the predicted logit for fine-grained event class $c \in \{1, \ldots, C\}$, and let $Y_{b,t,c} \in \{0,1\}$ be the corresponding ground-truth label. The fine-grained loss for labeled data is computed using binary cross-entropy:
\begin{equation}
\begin{split}
L_{\text{fine}}^{\text{lab}}
= \frac{1}{B \times T \times C}
\sum_{b=1}^{B} \sum_{t=1}^{T} \sum_{c=1}^{C}
\Big[
- Y_{b,t,c} \log \sigma(\hat{Y}_{b,t,c}) \\
- (1 - Y_{b,t,c}) \log \big(1 - \sigma(\hat{Y}_{b,t,c})\big)
\Big],
\end{split}
\end{equation}
where $\sigma(\cdot)$ denotes the sigmoid function.

\noindent\textbf{Unlabeled loss with pseudo-labeling.}
For unlabeled samples, pseudo-labels are generated using the model's current predictions. Let $\tilde{y}_{b,t}$ and $\tilde{Y}_{b,t,c}$ denote the pseudo-labels for the coarse-grained and fine-grained predictions, respectively. The unlabeled loss is defined analogously as
\begin{equation}
L_{\text{unlab}} = L_{\text{coarse}}^{\text{unlab}} + L_{\text{fine}}^{\text{unlab}}.
\end{equation}

\noindent\textbf{Total training objective.}
The overall loss for Stage~I is defined as
\begin{equation}
L_{\text{total}}
= L_{\text{lab}} + \lambda(e)\, L_{\text{unlab}},
\label{eq:stage1-loss}
\end{equation}
where
\begin{equation}
L_{\text{lab}} = L_{\text{coarse}}^{\text{lab}} + L_{\text{fine}}^{\text{lab}}.
\end{equation}
The weighting coefficient $\lambda(e)$ is scheduled based on the training epoch
$e$:
\begin{equation}
\lambda(e) =
\begin{cases}
0, & e < 30 \\[4pt]
\text{anneal}(0 \rightarrow 0.4), & 30 \le e < 90 \\[4pt]
0.4, & e \ge 90
\end{cases}
\end{equation}

\subsection{Multimodal Distillation}
\label{sec:MultimodalDistillation}

In Stage~II, we transfer the domain-adapted motion knowledge learned from skeletons to visual modalities, namely RGB and optical flow. The pretrained skeleton encoder from Stage~I serves as a teacher, while RGB and optical-flow encoders act as student networks.

For each modality $m \in \{\text{RGB}, \text{Flow}\}$, the student encoder projects the input $X_m$ into the same embedding space as the teacher:
\begin{equation}
F_m = \varepsilon^{m}_{\text{student}}(X_m).
\end{equation}

Knowledge distillation is achieved by minimizing the $\ell_2$ distance between student features and the teacher's pose embeddings:
\begin{equation}
L_m = \| F_{\text{pose}} - F_m \|_2.
\end{equation}

This stage follows standard multimodal distillation practices; further details can be found in related work such as \autocite{hatano2024multimodal,liu2026umegnet}. In the final stage, we train the temporal event detector on the target domain using the few-shot labeled dataset $\mathcal{D}_{\text{label}}$. The pretrained RGB and optical-flow student encoders obtained in Stage~II are retained, and the loss is computed in the same manner as in Stage~I to obtain the final model.

\noindent\textbf{MD-FED.}
MD-FED is a reduced version of AMD-FED that removes unlabeled data from Stage~I pretraining. It follows the same multimodal distillation pipeline but relies solely on labeled skeleton sequences when learning domain-specific pose representations. This variant isolates the effect of unlabeled data and annealed pseudo-labeling on downstream few-shot event detection. Preliminary experiments investigating attention mechanisms for this model are detailed in Appendix~\ref{subsec:attention_fusion}.

Implementation details of AMD-FED, including preprocessing, backbone selection, and training configuration, are provided in Appendix~\ref{appendix:b}.

\section{AWD: Adaptive Weight Distillation}
\label{sec:AWD}

In contrast to AMD-FED, which distills knowledge at the representation level, AWD adopts a \emph{prediction-layer distillation} strategy. Instead of aligning intermediate embeddings, the student network directly learns from the teacher's output predictions. To account for the teacher's reliability on unlabeled samples, AWD introduces a weighting term $W$ that scales the distillation loss adaptively based on the agreement between teacher and student:

\begin{equation}
W = \frac{1}{1 + p \, (d - 1)},
\end{equation}

where $p$ is a binary indicator representing whether the teacher's prediction is correct ($p=0$) or incorrect ($p=1$). Let $\hat{\mathbf{y}}^{(T)} \in \{0,1\}^C$ be the binarized teacher prediction, $\hat{\mathbf{y}}^{(S)} \in \{0,1\}^C$ the binarized student prediction, and $\mathbf{y} \in \{0,1\}^C$ the ground-truth fine label for a frame, where $C$ is the number of fine-grained classes. We define the distortion term $d$ as a mismatch-ratio statistic:
\begin{equation}
d =
\frac{
\sum_{c=1}^{C} \mathbf{1}\!\left[\hat{y}^{(T)}_c \neq \hat{y}^{(S)}_c\right] + 1
}{
\sum_{c=1}^{C} \mathbf{1}\!\left[\hat{y}^{(S)}_c \neq y_c\right] + 1
}.
\end{equation}
Here, the numerator measures the number of class-wise mismatches between teacher and student predictions, while the denominator measures the number of class-wise mismatches between the student prediction and the ground-truth label. The additive constant $1$ is used for smoothing to avoid division by zero. Thus, $d$ reflects how strongly the teacher-student disagreement compares to the student's actual error with respect to ground truth. When the teacher is correct, $W=1$ and the student fully follows the teacher. When the teacher's prediction is incorrect ($p=1$), the weight $W$ is reduced according to the ratio
\begin{equation}
\frac{1}{d}.
\end{equation}
If the student makes relatively fewer mismatches with respect to the ground-truth label than the level of disagreement between teacher and student, the ratio becomes large, which decreases $W$ and thus limits the influence of the unreliable teacher. For stability, $W$ is clipped to the range $[0,1]$.

In the original paper \autocite{iliopoulos2022weighted}, the weight parameters are estimated using a validation set via a nearest-neighbors regression approach. The validation samples provide a mapping between the teacher's and student's confidence scores and their corresponding reliability measures $(p,d)$. Each validation example contributes a pair of confidence values, one from the teacher and one from the student, as regression inputs, and a target pair consisting of the teacher's correctness $p$ and the observed distortion ratio $d$. During inference on unlabeled data, the same mapping is used to predict $p$ and $d$, enabling the model to assign adaptive weights based on the confidence and relative performance of the teacher. This approach mitigates error propagation from uncertain teacher predictions and enhances the robustness of the distillation process.

\subsection{Extension to Multi-Label Classification Setting}
\label{sec:AWDMultilabel}

The original paper designs the weighting mechanism for a multi-class classification task, where each sample belongs to exactly one class. In our setup, the action recognition problem is formulated as a multi-label classification task, where multiple labels can co-occur within a frame of a single clip. Therefore, we adapt the confidence estimation, validation mapping, and weight computation process accordingly.

\subsection{Averaged Confidence and Weight Estimation}

For each clip, we may have several frames, each containing a multi-labeled action such as \texttt{near\_return\_bh\_gs}. We obtain individual $p$ and $d$ values for each frame containing an action, as well as the confidences for the teacher and student models. We then average them at the clip level across the frames. The validation set is then constructed using input pairs

\begin{equation}
    (\bar{C}_s, \bar{C}_t)
\end{equation}

where $\bar{C}_s$ and $\bar{C}_t$ denote the averaged student and teacher confidences, respectively. The corresponding targets are the averaged $(\bar{p}, \bar{d})$. During inference, we compute the weight $W$ for a clip using the averaged confidence pair $(\bar{C}_s, \bar{C}_t)$ as the query to the $k$-Nearest Neighbours ($k$NN) model, which returns the averaged $(\bar{p}, \bar{d})$ from the validation mapping. The final weight is computed as

\begin{equation}
    W = \frac{1}{1 + \bar{p}(\bar{d} - 1)}, \quad \bar{p}, \bar{d} = kNN(\bar{C}_s, \bar{C}_t).
\end{equation}

\subsection{Confidence Computation for Multi-Label Predictions}

In the original multi-class setup, confidence is typically defined as the margin between the largest and second-largest softmax probabilities:

\begin{equation}
    C = \max(y) - 2^{\text{nd}}\max(y)
\end{equation}

For our multi-label setting, we define confidence within mutually exclusive label groups to reflect structured dependencies among action categories, as implemented in Listing~\ref{lst:postprocess}. The specific mapping of indices to classes is detailed in Table~\ref{tab:label_indices}. Specifically, labels $[0,1]$ form a binary pair where one is active and the other inactive, representing mutually exclusive states such as ``near'' and ``far.'' Similarly, labels $[2,3,4]$ represent a three-way exclusivity, where only one class within the group can be active at any time. The next set of groups, $[5,6]$ and $[7,8,\dots,12]$, are conditional on earlier decisions. Finally, label $[13]$ is treated as an independent binary indicator that does not depend on other groups. Within each group $g$, the confidence is computed as the difference between the largest and second-largest predicted probabilities, inter-class this time, while for label $[13]$, confidence is defined as the absolute difference between the two possible states,

\begin{equation}
    C_{13} = |y_{13} - (1 - y_{13})|.
\end{equation}

\begin{lstlisting}[language=Python, caption=Fine Label Post-processing Logic, label=lst:postprocess]
def fine_label_postprocess(pred):
    """
    Fine label post-processing logic for tennis pseudo labels.
    Ensures mutual exclusivity within label groups and rule-based
    hierarchy.
    """

    # [0,1]: near / far
    activate_one(pred, [0, 1])

    # [2,3,4]: serve / return / stroke
    activate_one(pred, [2, 3, 4])

    # if not serve
    if pred[2] != 1:
        # [5,6]: forehand / backhand
        activate_one(pred, [5, 6])
        # [7-12]: gs / slice / volley / smash / drop / lob
        activate_one(pred, [7, 8, 9, 10, 11, 12])

    # [13]: approach
    pred[13] = 1 if pred[13] >= 0.5 else 0

    return pred
\end{lstlisting}

\begin{table}[h]
\centering
\caption{Label Indices and Classes}
\label{tab:label_indices}
\begin{tabular}{|l|l|}
\hline
\textbf{Index} & \textbf{Class} \\ \hline
0  & near     \\ \hline
1  & far      \\ \hline
2  & serve    \\ \hline
3  & return   \\ \hline
4  & stroke   \\ \hline
5  & fh       \\ \hline
6  & bh       \\ \hline
7  & gs       \\ \hline
8  & slice    \\ \hline
9  & volley   \\ \hline
10 & smash    \\ \hline
11 & drop     \\ \hline
12 & lob      \\ \hline
13 & approach \\ \hline
\end{tabular}
\end{table}

The overall clip-level confidence is then obtained by averaging across all label groups,

\begin{equation}
    \bar{C} = \frac{1}{G} \sum_{g=1}^{G} C_g
\end{equation}

where $G$ is the number of label groups.

\chapter{Experimental Results}
\label{chap:ExperimentalResults}

This section presents the dataset, evaluation metrics, ablation study, and baseline comparison.

\section{Dataset}
\label{sec:Dataset}

We evaluate our method on two fine-grained sports datasets with diverse motion patterns. Each dataset follows a 3:1:1 split for training, validation, and testing, where training and validation samples come from the same videos, while test clips are sourced from separate videos.

\noindent\textbf{\boldmath $F^3$Set-Tennis.} $F^3$Set-Tennis \autocite{liu2025f3set} comprises 114 full-broadcast singles tennis matches featuring 75 professional players. It includes 11,584 rally-based clips (average 8.4s) and 42,846 shots across 1,000+ event types with precise hitting timestamps, offering multi-level granularity for analysis. Events are categorized into 8 sub-classes (e.g., player positioning, shot type, technique, direction, outcome). Since some sub-classes (e.g., shot direction and outcome) are not action-based, we focus on 5 pose-related sub-classes, denoted as $F^3$Set-Tennis(sub), with statistics shown in Table~\ref{tab:dataset-stats}.

\noindent\textbf{Figure Skating.} Figure Skating \autocite{hong2021video} consists of 11 videos (all 25 FPS) covering 371 short-program performances from the Winter Olympics (2010--2018) and World Championships (2017--2019). Following prior work, we refine the original annotations by manually re-labeling the take-off and landing frames of 10 jump and flying-spin classes, resulting in 20 event types (e.g., ``axel take-off'' and ``flip landing'').

\section{Evaluation Metrics}
\label{sec:EvaluationMetrics}

Following \autocite{liu2025f3set}, we assess our approach using two metrics that capture both temporal accuracy and classification performance: Edit score and mean F1 with temporal tolerance.

\noindent\textbf{(1) Edit Score} \autocite{lea2017tcn} measures the structural alignment between predicted and ground-truth event sequences using Levenshtein distance. It penalizes missing, redundant, and misordered predictions, making it suitable for tasks that require accurate event ordering and completeness.

\noindent\textbf{(2) Mean F1 Score with Temporal Tolerance} evaluates both classification and localization accuracy. A prediction is considered correct if it matches the event class and falls within a temporal window of $\delta$ frames around the ground-truth timestamp. We report the average F1 across all event categories, denoted as $F1_{\text{evt}}$. Unless otherwise specified, we adopt a strict tolerance of $\delta = 1$ frame.

\begin{table}[H]
\centering
\caption{Statistics of the $F^3$Set-Tennis(sub) dataset. Each sub-class contains multiple elements with corresponding event counts and proportions.}
\label{tab:dataset-stats}
\setlength{\tabcolsep}{6pt}
\renewcommand{\arraystretch}{1.2}
\begin{tabular}{c l r r}
\toprule
\textbf{Sub-Class} & \textbf{Element} & \textbf{Count} & \textbf{Proportion (\%)} \\
\midrule
\multirow{2}{*}{SC1}
 & Near & 21,467 & 50.1 \\
 & Far  & 21,362 & 49.9 \\
\midrule
\multirow{2}{*}{SC2}
 & Forehand & 27,802 & 64.9 \\
 & Backhand & 15,027 & 35.1 \\
\midrule
\multirow{3}{*}{SC3}
 & Serve  & 11,584 & 27.0 \\
 & Return & 8,216  & 19.2 \\
 & Stroke & 23,029 & 53.8 \\
\midrule
\multirow{6}{*}{SC4}
 & Ground stroke & 38,287 & 89.4 \\
 & Slice         & 3,358  & 7.8 \\
 & Volley        & 497    & 1.2 \\
 & Lob           & 334    & 0.8 \\
 & Drop          & 236    & 0.5 \\
 & Smash         & 117    & 0.3 \\
\midrule
\multirow{2}{*}{SC5}
 & Approach     & 964    & 2.3 \\
 & Non-approach & 41,865 & 97.7 \\
\bottomrule
\end{tabular}
\end{table}

\section{Ablation on Unlabeled Data Integration in Stage I}
\label{sec:AblationUnlabeledDataStageI}

Stage~I of AMD-FED adopts a semi-supervised training paradigm to pretrain the skeleton-based teacher encoder using both labeled and unlabeled data. While the overall objective in Eq.~\eqref{eq:stage1-loss} defines how unlabeled samples contribute through pseudo-labeling and an annealed loss weight, the manner in which unlabeled data is introduced during training can significantly affect the stability and quality of the learned representations. In this section, we conduct an ablation study to systematically examine different strategies for incorporating unlabeled samples in Stage~I.

\noindent\textbf{Experimental Protocol.}
We evaluate three unlabeled-data integration strategies inspired by pseudo-labeling-based semi-supervised learning \autocite{lee2013pseudo}:

\begin{enumerate}[topsep=0pt, itemsep=0pt, parsep=0pt]
\item \textbf{Joint training}, where labeled and unlabeled samples are used together from the first training epoch;
\item \textbf{Delayed integration}, in which the teacher encoder is trained solely on labeled data for the first $n$ epochs before unlabeled samples are introduced; and
\item \textbf{Best-model continuation}, where training with both labeled and unlabeled data resumes from the best checkpoint obtained during the labeled-only phase ($e \leq n$).
\end{enumerate}

For unlabeled samples, supervision is provided via pseudo-labels generated online from the model's current predictions. Since pseudo-labels are particularly noisy during early training, when the model is not yet sufficiently adapted to the target domain, directly incorporating unlabeled data can degrade learning. To mitigate this issue, the contribution of the unlabeled loss is modulated by the epoch-dependent weighting coefficient $\lambda(e)$, which gradually increases from zero to its target value as training progresses. This annealing strategy reduces the influence of unreliable pseudo-labels in early epochs and stabilizes optimization.

\begin{table}[t]
    \centering
    \footnotesize
    \caption{Ablation study on different strategies for incorporating unlabeled data in Stage~I of AMD-FED. ``From epoch 1'' indicates that unlabeled samples are used from the beginning of training, while ``After 30'' denotes delayed integration following 30 epochs of labeled-only training. The notation $0 \rightarrow \alpha$ ($e_s$--$e_e$) specifies the annealing schedule of the unlabeled loss weight $\lambda(e)$, which linearly increases from 0 to $\alpha$ between epochs $e_s$ and $e_e$. Best-model continuation yields the highest Edit score and is adopted in the final AMD-FED model.}
    \label{tab:stage1-ablation}
    \renewcommand{\arraystretch}{1.15}
    \begin{tabular}{l c}
        \toprule
        Setting & Edit (\%) \\
        \midrule
        Stage I (labeled only) & 75.79 \\
        \midrule
        \multicolumn{2}{l}{\textit{(a) Joint training}} \\
        From epoch 1, $0 \rightarrow 1$ (20--60) & 70.14 \\
        From epoch 1, $0 \rightarrow 1$ (20--80) & 75.12 \\
        \midrule
        \multicolumn{2}{l}{\textit{(b) Delayed integration}} \\
        After 30, $0 \rightarrow 3$ (30--60) & 74.26 \\
        After 30, $0 \rightarrow 1$ (30--60) & 76.43 \\
        After 30, $0 \rightarrow 1$ (30--90) & 77.55 \\
        After 30, $0 \rightarrow 0.7$ (30--90) & 78.26 \\
        After 30, $0 \rightarrow 0.4$ (30--90) & 78.66 \\
        \midrule
        \multicolumn{2}{l}{\textit{(c) Best-model continuation}} \\
        After 30, $0 \rightarrow 0.4$ (30--90) & \textbf{79.43} \\
        \bottomrule
    \end{tabular}
\end{table}

\noindent\textbf{Results and Analysis.}
The experimental results in Table~\ref{tab:stage1-ablation} show that \textbf{best-model continuation} consistently outperforms both joint training and delayed integration strategies. By initializing semi-supervised learning from the strongest labeled-only checkpoint, this strategy effectively reduces the impact of noisy pseudo-labels while still benefiting from abundant unlabeled data.

In contrast, \textbf{joint training} consistently performs worse than the labeled-only baseline. This degradation can be attributed to the training batch construction strategy: during dataset sampling, each mini-batch has a 50\% probability of being drawn from labeled or unlabeled data. Because the unlabeled loss weight $\lambda(e)$ is set to zero before epoch 20, the model effectively receives supervision from only half of the labeled data during early training. As a result, joint training under-utilizes labeled samples in the critical early optimization phase, leading to inferior representations compared to delayed integration and best-model continuation.

Moreover, adopting best-model continuation in Stage~I of AMD-FED leads to a clear performance improvement over labeled-only training. Across multiple 100-clip sets, the Edit score increases from $76.12 \pm 0.30$ to $79.51 \pm 0.18$. Note that Table~\ref{tab:stage1-ablation} reports an exploratory result on a single 100-clip set, while the reported averages summarize performance across several such sets. Based on these observations, AMD-FED adopts the best-model continuation strategy for incorporating unlabeled data in Stage~I.

\section{Baseline Comparisons}
\label{sec:BaselineComparisons}

Table~\ref{tab:edit_score} reports the Edit score and event-level F1 under the 50-clip and 100-clip settings.

First, \textbf{skeleton-based baselines consistently outperform RGB methods}. Under the 100-clip setting, all skeleton models achieve Edit scores above $76\%$, while the best RGB model reaches only $59.31\%$. A similar gap is observed in $F1_{evt}$, where skeleton models achieve $23\%$--$24\%$, compared to $11\%$--$15\%$ for RGB baselines. This highlights the robustness of pose representations under limited supervision.

Second, \textbf{RGB baselines remain comparatively unstable}. Although RegNet-Y backbones outperform ResNet-50 overall, their variance is noticeably larger, particularly in the 50-clip setting. Prior PES methods \textbf{F$^3$ED and T-DEED} also fail to close this modality gap, suggesting that improved temporal modeling alone is insufficient without robust representations. This suggests that appearance-driven models struggle to learn robust event semantics from limited supervision, even when equipped with temporal shift modules. The gap between RGB and skeleton baselines therefore justifies using the skeleton branch as the teacher modality in our framework.

Against these baselines, \textbf{our distillation-based methods provide substantial gains}. AWD improves the 100-clip Edit score to $77.03 \pm 0.81\%$, surpassing the best RGB baseline by $17.72$ points and slightly exceeding the strongest skeleton-only baseline. This shows that even prediction-level distillation can effectively transfer useful supervisory signals from the skeleton teacher to the student model. At the same time, the AWD configuration explored in this work should not be regarded as fully optimized. In particular, we did not conduct an ablation on the group-wise confidence design used for multi-label weighting, nor did we systematically compare against alternative prediction-distillation strategies from the literature. As such, AWD may admit further improvements through better confidence modeling or other prediction-level distillation methods that were beyond the scope of this study.

More importantly, \textbf{representation-level distillation is markedly more effective}. MD-FED raises the Edit score further to $81.89 \pm 1.14\%$, while AMD-FED achieves the best overall result of $83.84 \pm 0.96\%$ with an event F1 of $30.37 \pm 2.09\%$. Compared with AWD, AMD-FED improves Edit by $6.81$ points under the 100-clip setting, demonstrating the effectiveness of intermediate feature alignment. The additional gain of AMD-FED over MD-FED further demonstrates the value of incorporating unlabeled data through semi-supervised annealed pseudo-labeling.

Finally, \textbf{model capacity plays a critical role in few-shot settings}. Lower-capacity models such as ResNet-50 and RegNet-Y exhibit better generalization due to stronger inductive biases. In contrast, high-capacity architectures like SlowFast~\autocite{feichtenhofer2019slowfast} and VTN~\autocite{neimark2021video} tend to overfit, quickly driving training and validation losses to near zero. This suggests over-parameterization in low-data regimes, where models memorize dataset-specific patterns rather than learning transferable representations.

Overall, these results indicate that robust representation learning---rather than temporal modeling alone---is the key to effective few-shot event detection, and AMD-FED achieves the strongest generalization.

\begin{figure}[t]
    \centering
    \includegraphics[width=0.85\linewidth]{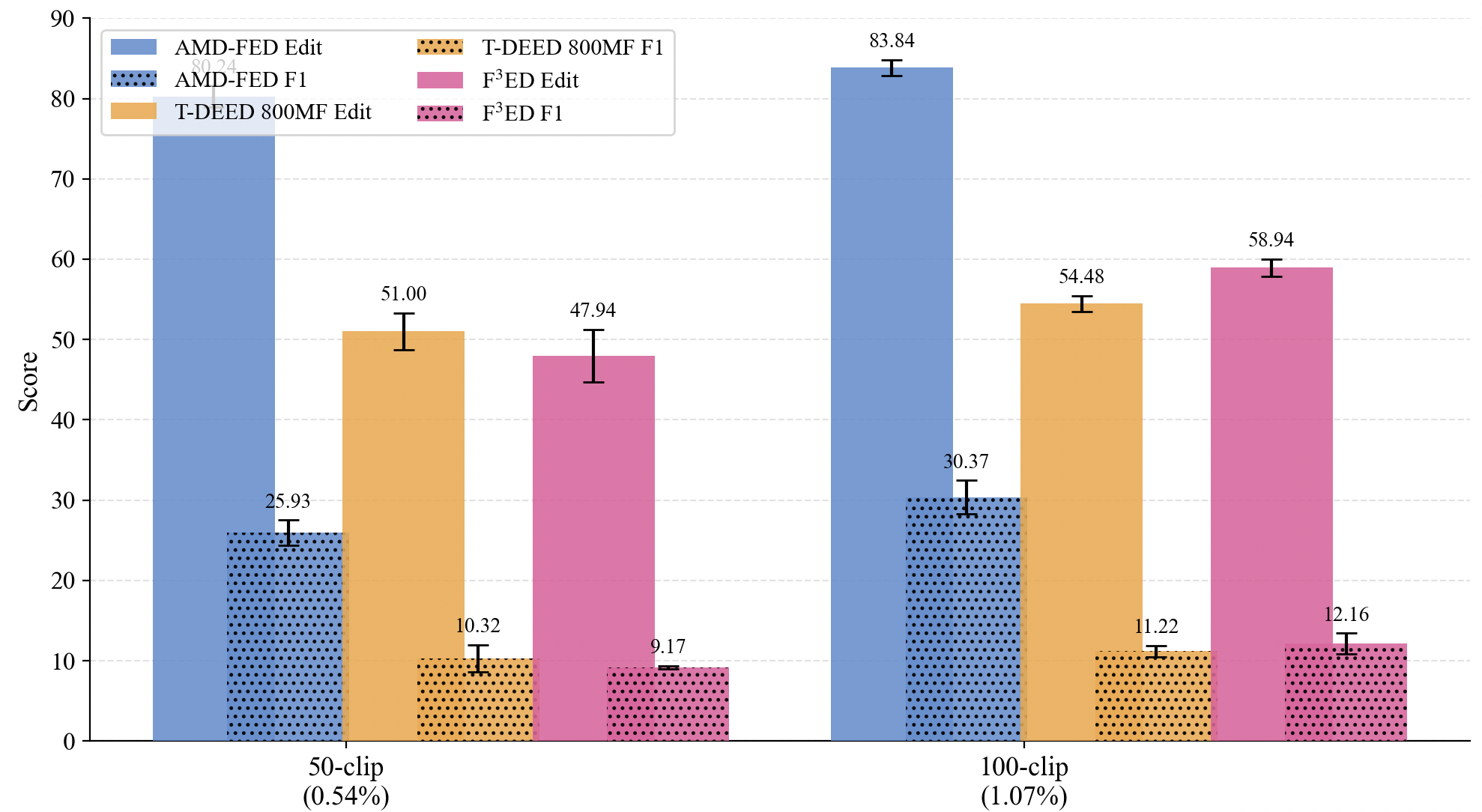}
    \caption{$F1_{\text{evt}}$ and Edit scores under few-shot ($k$-clip) training for the $F^3$Set-Tennis(sub) dataset. Percentages indicate the fraction of the full training set.}
    \label{fig:f3-comparison}
\end{figure}

\begin{figure}[t]
    \centering
    \includegraphics[width=0.85\linewidth]{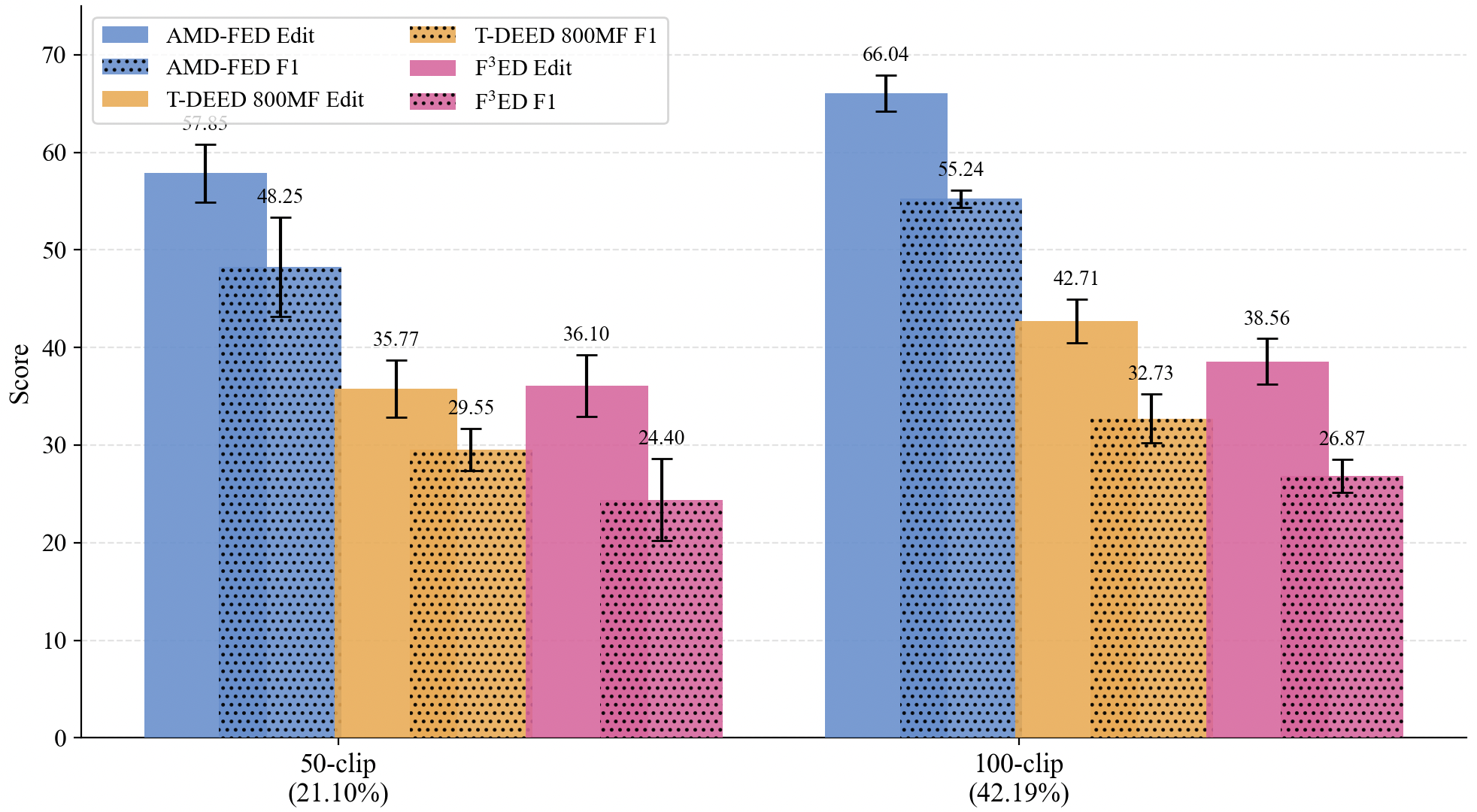}
    \caption{$F1_{\text{evt}}$ and Edit scores under few-shot ($k$-clip) training for the Figure Skating dataset. Percentages indicate the fraction of the full training set.}
    \label{fig:fs-comparison}
\end{figure}

Figures~\ref{fig:f3-comparison} and~\ref{fig:fs-comparison} provide a side-by-side comparison across the two datasets. On $F^3$Set-Tennis(sub), AMD-FED maintains the strongest overall performance as the number of training clips varies. Given that AMD-FED achieves the strongest performance on tennis, we further examine its robustness on the Figure Skating dataset. Under the 100-clip setting, AMD-FED achieves an Edit score of $66.04 \pm 1.87$ and an event-level F1 of $55.24 \pm 0.90$, substantially outperforming the strongest baseline, T-DEED$_{\text{800MF}}$, which obtains $42.71 \pm 2.22$ Edit and $32.73 \pm 2.53$ F1. In the 50-clip regime, AMD-FED reaches $57.85 \pm 2.95$ Edit and $48.25 \pm 5.09$ F1, again clearly exceeding the strongest baseline, F$^3$ED, which achieves $36.10 \pm 3.15$ Edit and $24.40 \pm 4.22$ F1. These results show that AMD-FED consistently outperforms the best competing baseline under both few-shot settings. Although this comparison is selective rather than exhaustive, it suggests that the benefit of AMD-FED is not limited to tennis, but also transfers to a structurally different fine-grained sports setting with distinct motion patterns and event definitions.

\begin{table}[t]
    \centering
    \footnotesize
    \caption{Edit score (\%) and event-level $F1_{\text{evt}}$ (mean $\pm$ standard deviation) on the $F^3$Set-Tennis(sub) dataset under 50-clip and 100-clip few-shot settings.}
    \label{tab:edit_score}
    \setlength{\tabcolsep}{5pt}
    \renewcommand{\arraystretch}{1.15}
    \begin{tabular}{p{4.2cm} cc cc}
        \toprule
        \multirow{2}{*}{Method} & \multicolumn{2}{c}{50-clip} & \multicolumn{2}{c}{100-clip} \\
        \cmidrule(lr){2-3} \cmidrule(lr){4-5}
        & Edit & $F1_{\text{evt}}$ & Edit & $F1_{\text{evt}}$ \\
        \midrule
        \multicolumn{5}{l}{\textit{(a) RGB modality}} \\
        RegNet-Y-200MF (TSM)~\autocite{radosavovic2020designing,lin2019tsm}
        & $47.90 \pm 2.12$ & $9.06 \pm 0.26$ & $55.35 \pm 1.19$ & $12.31 \pm 1.34$ \\
        RegNet-Y-800MF (TSM)~\autocite{radosavovic2020designing,lin2019tsm}
        & $47.11 \pm 9.42$ & $8.83 \pm 2.22$ & $59.31 \pm 1.84$ & $14.93 \pm 1.81$ \\
        ResNet-50 (TSM)~\autocite{he2016deep,lin2019tsm}
        & $49.29 \pm 1.84$ & $8.37 \pm 0.33$ & $52.69 \pm 4.46$ & $11.67 \pm 1.56$ \\
        \midrule
        \multicolumn{5}{l}{\textit{(b) 2D skeleton modality}} \\
        STGCN~\autocite{yan2018stgcn} & $69.39 \pm 2.58$ & $19.24 \pm 1.49$ & $76.29 \pm 0.46$ & $24.16 \pm 0.80$ \\
        MSG3D~\autocite{liu2020disentangling} & $71.07 \pm 3.13$ & $21.97 \pm 1.76$ & $76.65 \pm 1.00$ & $23.17 \pm 3.06$ \\
        AAGCN~\autocite{shi2020skeleton} & $68.93 \pm 0.89$ & $21.63 \pm 1.75$ & $76.19 \pm 2.12$ & $24.22 \pm 2.86$ \\
        CTRGCN~\autocite{chen2021channel} & $69.56 \pm 2.59$ & $20.26 \pm 0.46$ & $76.58 \pm 1.35$ & $23.59 \pm 1.18$ \\
        STGCN++~\autocite{duan2022pyskl} & $68.17 \pm 0.83$ & $19.48 \pm 1.38$ & $76.12 \pm 0.30$ & $24.30 \pm 1.24$ \\
        \midrule
        \multicolumn{5}{l}{\textit{(c) Prior PES methods}} \\
        F$^3$ED (contextualized)~\autocite{liu2025f3set} & $47.94 \pm 3.28$ & $9.17 \pm 0.22$ & $58.94 \pm 1.06$ & $12.16 \pm 1.28$ \\
        T-DEED$_{\text{200MF}}$~\autocite{xarles2024tdeed} & $38.67 \pm 9.05$ & $6.45 \pm 1.04$ & $49.72 \pm 1.55$ & $10.17 \pm 0.87$ \\
        T-DEED$_{\text{800MF}}$~\autocite{xarles2024tdeed} & $51.00 \pm 2.27$ & $10.32 \pm 1.67$ & $54.48 \pm 0.99$ & $11.22 \pm 0.70$ \\
        \midrule
        \multicolumn{5}{l}{\textit{(d) Our approach}} \\
        \makecell[l]{AWD \\ (Pred-Distillation)}
        & $72.58 \pm 5.12$ & $18.47 \pm 2.89$ & $77.03 \pm 0.81$ & $20.52 \pm 0.67$ \\
        \makecell[l]{MD-FED \\ (Repr-Distillation)}
        & $77.91 \pm 1.67$ & $25.13 \pm 2.10$ & $81.89 \pm 1.14$ & $29.76 \pm 1.21$ \\
        \makecell[l]{\textbf{AMD-FED} \\ (Repr-Distillation)}
        & $\mathbf{80.24 \pm 1.35}$ & $\mathbf{25.93 \pm 1.59}$ & $\mathbf{83.84 \pm 0.96}$ & $\mathbf{30.37 \pm 2.09}$ \\
        \bottomrule
    \end{tabular}
\end{table}

\clearpage

\chapter{Conclusion}

\section{Conclusion and Future Work}
\label{sec:ConclusionAndFutureWork}

In this paper, we studied few-shot Precise Event Spotting (PES) under limited supervision and proposed two distillation-based frameworks for this setting: \textbf{AWD}, which transfers knowledge at the prediction level, and \textbf{AMD-FED}, which performs representation-level multimodal distillation with semi-supervised annealed pseudo-labeling. By leveraging the robustness of skeleton-based teachers and transferring domain-adapted motion knowledge to visual modalities, our approach improves the ability of appearance-based models to generalize in low-data regimes.

Extensive experiments on \mbox{$F^3$Set-Tennis(sub)} show that both distillation strategies consistently improve over conventional RGB-based baselines in few-shot scenarios, while \textbf{AMD-FED} achieves the strongest overall performance. The gains over both visual and skeleton-based baselines indicate that representation-level transfer is especially effective for fine-grained event spotting, where subtle motion cues and precise timing are critical. Motivated by this result, we further validate AMD-FED on the Figure Skating dataset and observe a similar advantage over a selective baseline comparison, suggesting that the robustness of the representation-level framework extends beyond tennis to sports with different motion structures and event definitions. At the same time, the Figure Skating study should be interpreted as a brief validation rather than a comprehensive benchmark, since we compared against only a selective set of baselines rather than an exhaustive list of competing methods.

Overall, these findings confirm that multimodal distillation, together with semi-supervised teacher adaptation, is a promising direction for addressing annotation scarcity and improving generalization in few-shot precise event spotting. For future work, we aim to evaluate the proposed framework on additional sports datasets and investigate stronger semi-supervised and teacher-adaptation strategies for handling noisy pseudo-labels.

\vspace{5mm}
\noindent\textbf{Publication Note.} A shortened version of this report has been accepted for presentation at ISACE 2026, the 3rd International Sports Analytics Conference and Exhibition.

\clearpage
\printbibliography[heading=bibintoc, title={References}]

@article{javed2016efficient,
  title={An efficient framework for automatic highlights generation from sports videos},
  author={Javed, A. and Bajwa, KB and Malik, H. and Irtaza, A.},
  journal={IEEE Signal Processing Letters},
  volume={23},
  number={7},
  pages={954--958},
  year={2016},
  publisher={IEEE}
}

@article{chen2024interpretable,
  author       = {Chen, G.},
  title        = {RETRACTED ARTICLE: An interpretable composite CNN and GRU for fine-grained martial arts motion modeling using big data analytics and machine learning},
  journal      = {Soft Computing (Berlin, Germany)},
  volume       = {28},
  number       = {3},
  pages        = {2223--2243},
  year         = {2024},
  note         = {Retracted},
  doi          = {10.1007/s00500-023-09565-z}
}

@inproceedings{liu2023insight,
  author    = {Liu, Zhaoyu and Jiang, Kan and Hou, Zhe and Lin, Yun and Dong, Jin Song},
  title     = {Insight Analysis for Tennis Strategy and Tactics},
  booktitle = {2023 IEEE International Conference on Data Mining (ICDM)},
  pages     = {1169--1174},
  year      = {2023},
  organization = {IEEE},
  doi       = {10.1109/ICDM58522.2023.00143},
}

@article{ma2024refereeing,
  author  = {Ma, E. and Kabala, Z. J.},
  title   = {Refereeing the Sport of Squash with a Machine Learning System},
  journal = {Machine Learning and Knowledge Extraction},
  volume  = {6},
  number  = {1},
  pages   = {506--553},
  year    = {2024}
}

@inproceedings{decroos2018automatic,
  author    = {Decroos, Tom and Van Haaren, Jan and Davis, Jesse},
  title     = {Automatic Discovery of Tactics in Spatio-Temporal Soccer Match Data},
  booktitle = {Proceedings of the 24th ACM SIGKDD International Conference on Knowledge Discovery \& Data Mining},
  pages     = {223--232},
  year      = {2018},
  publisher = {ACM}
}

@article{antonini2024engineering,
  author  = {Antonini, V. and Mileo, A. and Roantree, M.},
  title   = {Engineering Features from Raw Sensor Data to Analyse Player Movements during Competition},
  journal = {Sensors},
  volume  = {24},
  number  = {4},
  pages   = {1308},
  year    = {2024}
}

@article{black2016monitoring,
  author  = {Black, G. M. and Gabbett, T. J. and Cole, M. H. and Naughton, G.},
  title   = {Monitoring Workload in Throwing-Dominant Sports: A Systematic Review},
  journal = {Sports Medicine},
  volume  = {46},
  number  = {10},
  pages   = {1503--1516},
  year    = {2016}
}

@article{bian2024p2anet,
  author    = {Bian, Jiang and Li, Xuhong and Wang, Tao and Wang, Qingzhong and Huang, Jun and Liu, Chen and Zhao, Jun and Lu, Feixiang and Dou, Dejing and Xiong, Haoyi},
  title     = {P2ANet: A Large-Scale Benchmark for Dense Action Detection from Table Tennis Match Broadcasting Videos},
  journal   = {ACM Transactions on Multimedia Computing, Communications, and Applications},
  volume    = {20},
  number    = {4},
  pages     = {1--23},
  year      = {2024},
  publisher = {ACM}
}

@inproceedings{liu2025f3set,
  author    = {Liu, Zhaoyu and Jiang, Kan and Ma, Murong and Hou, Zhe and Lin, Yun and Dong, Jin Song},
  title     = {F$^3$Set: Towards Analyzing Fast, Frequent, and Fine-Grained Events from Videos},
  booktitle = {International Conference on Learning Representations (ICLR)},
  year      = {2025}
}

@inproceedings{shao2020finegym,
  author    = {Shao, Dian and Zhao, Yue and Dai, Bo and Lin, Dahua},
  title     = {FineGym: A Hierarchical Video Dataset for Fine-Grained Action Understanding},
  booktitle = {Proceedings of the IEEE/CVF Conference on Computer Vision and Pattern Recognition (CVPR)},
  pages     = {2616--2625},
  year      = {2020}
}

@inproceedings{wang2023shuttleset,
  author    = {Wang, Wei-Yao and Huang, Yung-Chang and Ik, Tsi-Ui and Peng, Wen-Chih},
  title     = {ShuttleSet: A Human-Annotated Stroke-Level Singles Dataset for Badminton Tactical Analysis},
  booktitle = {Proceedings of the 29th ACM SIGKDD Conference on Knowledge Discovery and Data Mining (KDD)},
  pages     = {5126--5136},
  year      = {2023},
  publisher = {ACM}
}

@inproceedings{xu2022finediving,
  author    = {Xu, Jinglin and Rao, Yongming and Yu, Xumin and Chen, Guangyi and Zhou, Jie and Lu, Jiwen},
  title     = {FineDiving: A Fine-Grained Dataset for Procedure-Aware Action Quality Assessment},
  booktitle = {Proceedings of the IEEE/CVF Conference on Computer Vision and Pattern Recognition (CVPR)},
  pages     = {2949--2958},
  year      = {2022}
}

@inproceedings{hong2022spotting,
  author    = {Hong, James and Zhang, Haotian and Gharbi, Michaël and Fisher, Matthew and Fatahalian, Kayvon},
  title     = {Spotting Temporally Precise, Fine-Grained Events in Video},
  booktitle = {European Conference on Computer Vision (ECCV)},
  pages     = {33--51},
  year      = {2022},
  publisher = {Springer}
}

@inproceedings{liu2026umegnet,
  author    = {Liu, Zhaoyu and Jiang, Kan and Ma, Murong and Hou, Zhe and Lin, Yun and Dong, Jin Song},
  title     = {Few-Shot Precise Event Spotting via Unified Multi-Entity Graph and Distillation},
  booktitle = {Proceedings of the 40th AAAI Conference on Artificial Intelligence (AAAI)},
  year      = {2026},
  url       = {https://github.com/LZYAndy/UMEG-Net}
}

@inproceedings{liu2020disentangling,
  author    = {Liu, Ziyu and Zhang, Hongwen and Chen, Zhenghao and Wang, Zhiyong and Ouyang, Wanli},
  title     = {Disentangling and Unifying Graph Convolutions for Skeleton-Based Action Recognition},
  booktitle = {Proceedings of the IEEE/CVF Conference on Computer Vision and Pattern Recognition (CVPR)},
  year      = {2020},
  pages     = {140--149},
  doi       = {10.1109/CVPR42600.2020.00022},
  url       = {https://openaccess.thecvf.com/content_CVPR_2020/html/Liu_Disentangling_and_Unifying_Graph_Convolutions_for_Skeleton-Based_Action_Recognition_CVPR_2020_paper.html}
}

@inproceedings{sun2019deep,
  author    = {Sun, Ke and Xiao, Bin and Liu, Dong and Wang, Jingdong},
  title     = {Deep High-Resolution Representation Learning for Human Pose Estimation},
  booktitle = {Proceedings of the IEEE/CVF Conference on Computer Vision and Pattern Recognition (CVPR)},
  year      = {2019}
}

@misc{wu2019detectron2,
  author       = {Wu, Yuxin and Kirillov, Alexander and Massa, Francisco and Lo, Wan-Yen and Girshick, Ross},
  title        = {Detectron2},
  howpublished = {\url{https://github.com/facebookresearch/detectron2}},
  year         = {2019}
}

@inproceedings{carreira2017quo,
  author    = {Carreira, Joao and Zisserman, Andrew},
  title     = {Quo Vadis, Action Recognition? A New Model and the Kinetics Dataset},
  booktitle = {Proceedings of the IEEE Conference on Computer Vision and Pattern Recognition (CVPR)},
  year      = {2017},
  pages     = {6299--6308},
  doi       = {10.1109/CVPR.2017.671},
  url       = {https://openaccess.thecvf.com/content_cvpr_2017/html/Carreira_Quo_Vadis_Action_CVPR_2017_paper.html}
}

@inproceedings{feichtenhofer2019slowfast,
  author    = {Feichtenhofer, Christoph and Fan, Haoqi and Malik, Jitendra and He, Kaiming},
  title     = {SlowFast Networks for Video Recognition},
  booktitle = {Proceedings of the IEEE/CVF International Conference on Computer Vision (ICCV)},
  pages     = {6202--6211},
  year      = {2019}
}

@inproceedings{lin2019tsm,
  author    = {Lin, Ji and Gan, Chuang and Han, Song},
  title     = {TSM: Temporal Shift Module for Efficient Video Understanding},
  booktitle = {Proceedings of the IEEE/CVF International Conference on Computer Vision (ICCV)},
  pages     = {7083--7093},
  year      = {2019}
}

@article{wang2018temporal,
  author  = {Wang, Limin and Xiong, Yuanjun and Wang, Zhe and Qiao, Yu and Lin, Dahua and Tang, Xiaoou and Van Gool, Luc},
  title   = {Temporal Segment Networks for Action Recognition in Videos},
  journal = {IEEE Transactions on Pattern Analysis and Machine Intelligence},
  volume  = {41},
  number  = {11},
  pages   = {2740--2755},
  year    = {2018}
}

@article{fei2006one,
  author  = {Fei-Fei, Li and Fergus, Rob and Perona, Pietro},
  title   = {One-Shot Learning of Object Categories},
  journal = {IEEE Transactions on Pattern Analysis and Machine Intelligence},
  volume  = {28},
  number  = {4},
  pages   = {594--611},
  year    = {2006},
  month   = {apr}
}

@article{salakhutdinov2013learning,
  author  = {Salakhutdinov, Ruslan and Tenenbaum, Joshua B. and Torralba, Antonio},
  title   = {Learning with Hierarchical-Deep Models},
  journal = {IEEE Transactions on Pattern Analysis and Machine Intelligence},
  volume  = {35},
  number  = {8},
  pages   = {1958--1971},
  year    = {2013},
  month   = {aug}
}

@inproceedings{finn2017maml,
  author    = {Finn, Chelsea and Abbeel, Pieter and Levine, Sergey},
  title     = {Model-Agnostic Meta-Learning for Fast Adaptation of Deep Networks},
  booktitle = {Proceedings of the 34th International Conference on Machine Learning (ICML)},
  year      = {2017}
}

@inproceedings{hinton1987fast,
  author    = {Hinton, Geoffrey E. and Plaut, David C.},
  title     = {Using Fast Weights to Deblur Old Memories},
  booktitle = {Proceedings of the Ninth Annual Conference of the Cognitive Science Society (CogSci)},
  year      = {1987}
}

@incollection{thrun1998learning,
  author    = {Thrun, Sebastian and Pratt, Lorien},
  title     = {Learning to Learn: Introduction and Overview},
  booktitle = {Learning to Learn},
  pages     = {3--17},
  year      = {1998},
  publisher = {Springer}
}

@inproceedings{snell2017prototypical,
  author    = {Snell, Jake and Swersky, Kevin and Zemel, Richard S.},
  title     = {Prototypical Networks for Few-Shot Learning},
  booktitle = {Advances in Neural Information Processing Systems (NeurIPS)},
  year      = {2017}
}

@inproceedings{sung2018learning,
  author    = {Sung, Flood and Yang, Yongxin and Zhang, Li and Xiang, Tao and Torr, Philip H. S. and Hospedales, Timothy M.},
  title     = {Learning to Compare: Relation Network for Few-Shot Learning},
  booktitle = {Proceedings of the IEEE Conference on Computer Vision and Pattern Recognition (CVPR)},
  year      = {2018}
}

@inproceedings{vinyals2016matching,
  author    = {Vinyals, Oriol and Blundell, Charles and Lillicrap, Tim and Kavukcuoglu, Koray and Wierstra, Daan},
  title     = {Matching Networks for One-Shot Learning},
  booktitle = {Advances in Neural Information Processing Systems (NeurIPS)},
  year      = {2016}
}

@inproceedings{mishra2018snail,
  author    = {Mishra, Nikhil and Rohaninejad, Mohammad and Chen, Xiang and Abbeel, Pieter},
  title     = {SNAIL: A Simple Neural Attentive Meta-Learner},
  booktitle = {International Conference on Learning Representations (ICLR)},
  year      = {2018}
}

@inproceedings{munkhdalai2017metanetworks,
  author    = {Munkhdalai, Tsendsuren and Yu, Hong},
  title     = {Meta Networks},
  booktitle = {Proceedings of the 34th International Conference on Machine Learning (ICML)},
  year      = {2017}
}

@inproceedings{oreshkin2018tadam,
  author    = {Oreshkin, Boris N. and Rodríguez, Pau and Lacoste, Alexandre},
  title     = {TADAM: Task Dependent Adaptive Metric for Improved Few-Shot Learning},
  booktitle = {Advances in Neural Information Processing Systems (NeurIPS)},
  year      = {2018}
}

@inproceedings{santoro2016meta,
  author    = {Santoro, Adam and Bartunov, Sergey and Botvinick, Matthew and Wierstra, Daan and Lillicrap, Timothy P.},
  title     = {Meta-Learning with Memory-Augmented Neural Networks},
  booktitle = {Proceedings of the 33rd International Conference on Machine Learning (ICML)},
  year      = {2016}
}

@inproceedings{ravi2017optimization,
  author    = {Ravi, Sachin and Larochelle, Hugo},
  title     = {Optimization as a Model for Few-Shot Learning},
  booktitle = {International Conference on Learning Representations (ICLR)},
  year      = {2017}
}

@inproceedings{lee2018gradient,
  author    = {Lee, Yang and Choi, Sang},
  title     = {Gradient-Based Meta-Learning with Learned Layerwise Metric and Subspace},
  booktitle = {Proceedings of the 35th International Conference on Machine Learning (ICML)},
  year      = {2018}
}

@inproceedings{grant2018recasting,
  author    = {Grant, Elliott and Finn, Chelsea and Levine, Sergey and Darrell, Trevor and Griffiths, Thomas L.},
  title     = {Recasting Gradient-Based Meta-Learning as Hierarchical Bayes},
  booktitle = {International Conference on Learning Representations (ICLR)},
  year      = {2018}
}

@inproceedings{zhang2018metagan,
  author    = {Zhang, Rui and Che, Tongzheng and Graham, Zoubin and Bengio, Yoshua and Song, Yang},
  title     = {MetaGAN: An Adversarial Approach to Few-Shot Learning},
  booktitle = {Advances in Neural Information Processing Systems (NeurIPS)},
  year      = {2018}
}

@inproceedings{yang2018one,
  author    = {Yang, Hongtao and He, Xuming and Porikli, Fatih},
  title     = {One-Shot Action Localization by Learning Sequence Matching Network},
  booktitle = {Proceedings of the IEEE Conference on Computer Vision and Pattern Recognition (CVPR)},
  pages     = {1450--1459},
  year      = {2018}
}

@article{nag2021few,
  author  = {Nag, Sauradip and Zhu, Xiatian and Xiang, Tao},
  title   = {Few-Shot Temporal Action Localization with Query Adaptive Transformer},
  journal = {arXiv preprint arXiv:2110.10552},
  year    = {2021}
}

@inproceedings{hong2021video,
  author    = {Hong, James and Fisher, Matthew and Gharbi, Micha{\"e}l and Fatahalian, Kayvon},
  title     = {Video Pose Distillation for Few-Shot, Fine-Grained Sports Action Recognition},
  booktitle = {Proceedings of the IEEE/CVF International Conference on Computer Vision (ICCV)},
  year      = {2021}
}

@article{hinton2015distilling,
  author  = {Hinton, Geoffrey E. and Vinyals, Oriol and Dean, Jeff},
  title   = {Distilling the Knowledge in a Neural Network},
  journal = {CoRR},
  volume  = {abs/1503.02531},
  year    = {2015},
  url     = {https://arxiv.org/abs/1503.02531}
}

@inproceedings{ahn2019variational,
  author    = {Ahn, Sungsoo and Xu, Song},
  title     = {Variational Information Distillation for Knowledge Transfer},
  booktitle = {Proceedings of the IEEE/CVF Conference on Computer Vision and Pattern Recognition (CVPR)},
  pages     = {9163--9171},
  year      = {2019}
}

@article{chen2020learning,
  author  = {Chen, Hanting and Wang, Yao},
  title   = {Learning Student Networks via Feature Embedding},
  journal = {IEEE Transactions on Neural Networks and Learning Systems},
  pages   = {25--35},
  year    = {2020}
}

@inproceedings{stanton2021knowledge,
  author    = {Stanton, Samuel and Iscen, Pinar},
  title     = {Does Knowledge Distillation Really Work?},
  booktitle = {Advances in Neural Information Processing Systems (NeurIPS)},
  pages     = {6906--6919},
  year      = {2021}
}

@article{dehghani2017fidelity,
  author  = {Dehghani, Mostafa and Mehrasa, Arash},
  title   = {Fidelity-Weighted Learning},
  journal = {arXiv preprint arXiv:1711.02799},
  year    = {2017},
  url     = {https://arxiv.org/abs/1711.02799}
}

@article{lang2022training,
  author  = {Lang, Hunter and Vlachos, Andreas},
  title   = {Training Subset Selection for Weak Supervision},
  journal = {arXiv preprint arXiv:2206.02914},
  year    = {2022},
  url     = {https://arxiv.org/abs/2206.02914}
}

@inproceedings{iliopoulos2022weighted,
  author    = {Iliopoulos, Fotis and Karypis, Vassilis},
  title     = {Weighted Distillation with Unlabeled Examples},
  booktitle = {Advances in Neural Information Processing Systems (NeurIPS)},
  year      = {2022}
}

@inproceedings{duan2022pyskl,
  author    = {Duan, Haodong and Wang, Jiaqi and Chen, Kai and Lin, Dahua},
  title     = {PySKL: Towards Good Practices for Skeleton Action Recognition},
  booktitle = {Proceedings of the 30th ACM International Conference on Multimedia},
  pages     = {7351--7354},
  year      = {2022},
  publisher = {ACM}
}

@inproceedings{chen2021channel,
  author    = {Chen, Yuxin and Zhang, Ziqi and Yuan, Chunfeng and Li, Bing and Deng, Ying and Hu, Weiming},
  title     = {Channel-Wise Topology Refinement Graph Convolution for Skeleton-Based Action Recognition},
  booktitle = {Proceedings of the IEEE/CVF International Conference on Computer Vision (ICCV)},
  pages     = {13359--13368},
  year      = {2021}
}

@article{shi2020skeleton,
  author  = {Shi, Lei and Zhang, Yifan and Cheng, Jian and Lu, Hanqing},
  title   = {Skeleton-Based Action Recognition with Multi-Stream Adaptive Graph Convolutional Networks},
  journal = {IEEE Transactions on Image Processing},
  volume  = {29},
  pages   = {9532--9545},
  year    = {2020}
}

@inproceedings{yan2018stgcn,
  author    = {Yan, Sijie and Xiong, Yuanjun and Lin, Dahua},
  title     = {Spatial Temporal Graph Convolutional Networks for Skeleton-Based Action Recognition},
  booktitle = {Proceedings of the AAAI Conference on Artificial Intelligence (AAAI)},
  year      = {2018}
}

@inproceedings{radosavovic2020designing,
  author    = {Radosavovic, Ilija and Kosaraju, Raj Prateek and Girshick, Ross and He, Kaiming and Doll{\'a}r, Piotr},
  title     = {Designing Network Design Spaces},
  booktitle = {Proceedings of the IEEE/CVF Conference on Computer Vision and Pattern Recognition (CVPR)},
  pages     = {10428--10436},
  year      = {2020}
}

@inproceedings{he2016deep,
  author    = {He, Kaiming and Zhang, Xiangyu and Ren, Shaoqing and Sun, Jian},
  title     = {Deep Residual Learning for Image Recognition},
  booktitle = {Proceedings of the IEEE Conference on Computer Vision and Pattern Recognition (CVPR)},
  pages     = {770--778},
  year      = {2016}
}

@inproceedings{lee2013pseudo,
  author    = {Lee, Dong-Hyun},
  title     = {Pseudo-Label: The Simple and Efficient Semi-Supervised Learning Method for Deep Neural Networks},
  booktitle = {Workshop on Challenges in Representation Learning, ICML},
  year      = {2013}
}

@inproceedings{neimark2021video,
  author    = {Neimark, Daniel and Bar, Ofer and Zohar, Matan and Asselmann, Dotan},
  title     = {Video Transformer Network},
  booktitle = {Proceedings of the IEEE/CVF Conference on Computer Vision and Pattern Recognition (CVPR)},
  pages     = {3163--3172},
  year      = {2021}
}

@inproceedings{hatano2024multimodal,
  author    = {Hatano, Masashi and Hachiuma, Ryo and Fujii, Ryo and Saito, Hideo},
  title     = {Multimodal Cross-Domain Few-Shot Learning for Egocentric Action Recognition},
  booktitle = {Proceedings of the European Conference on Computer Vision (ECCV)},
  year      = {2024}
}

@techreport{krizhevsky2009learning,
  author      = {Krizhevsky, Alex},
  title       = {Learning Multiple Layers of Features from Tiny Images},
  institution = {University of Toronto},
  year        = {2009}
}

@misc{lecun2010mnist,
  author       = {LeCun, Yann and Cortes, Corinna and Burges, C. J.},
  title        = {MNIST Handwritten Digit Database},
  howpublished = {\url{http://yann.lecun.com/exdb/mnist/}},
  year         = {2010}
}

@misc{chen2026action100m,
  author       = {Chen, Delong and Kasarla, Tejaswi and Bang, Yejin and Shukor, Mustafa and Chung, Willy and Yu, Jade and Bolourchi, Allen and Moutakanni, Théo and Fung, Pascale},
  title        = {Action100M: A Large-scale Video Action Dataset},
  year         = {2026},
  eprint       = {2601.10592},
  archivePrefix= {arXiv},
  primaryClass = {cs.CV},
  url          = {https://arxiv.org/abs/2601.10592}
}

@misc{du2023semisupervisedlearningweightawaredistillation,
  author       = {Du, Pan and Zhao, Suyun and Sheng, Zisen and Li, Cuiping and Chen, Hong},
  title        = {Semi-Supervised Learning via Weight-aware Distillation under Class Distribution Mismatch},
  year         = {2023},
  eprint       = {2308.11874},
  archivePrefix= {arXiv},
  primaryClass = {cs.CV},
  url          = {https://arxiv.org/abs/2308.11874}
}

@misc{kontonis2023slamstudentlabelmixingdistillation,
  author       = {Kontonis, Vasilis and Iliopoulos, Fotis and Trinh, Khoa and Baykal, Cenk and Menghani, Gaurav and Vee, Erik},
  title        = {SLaM: Student-Label Mixing for Distillation with Unlabeled Examples},
  year         = {2023},
  eprint       = {2302.03806},
  archivePrefix= {arXiv},
  primaryClass = {cs.LG},
  url          = {https://arxiv.org/abs/2302.03806}
}

@misc{bock2023wear,
  author       = {Bock, M. and Moeller, M. and Van Laerhoven, K. and Kuehne, H.},
  title        = {Wear: A Multimodal Dataset for Wearable and Egocentric Video Activity Recognition},
  year         = {2023},
  eprint       = {2304.05088},
  archivePrefix= {arXiv},
  primaryClass = {cs.CV},
  url          = {https://arxiv.org/abs/2304.05088}
}

@inproceedings{damen2018epic,
  author    = {Damen, D. and Doughty, H. and Farinella, G. M. and Fidler, S. and Furnari, A. and Kazakos, E. and Moltisanti, D. and Munro, J. and Perrett, T. and Price, W. and Wray, M.},
  title     = {Scaling Egocentric Vision: The EPIC-Kitchens Dataset},
  booktitle = {European Conference on Computer Vision (ECCV)},
  year      = {2018}
}

@inproceedings{grauman2022ego4d,
  author    = {Grauman, K. and Westbury, A. and Byrne, E. and Chavis, Z. and Furnari, A. and Girdhar, R. and Hamburger, J. and Jiang, H. and Liu, M. and Liu, X. and Martin, M. and Nagarajan, T. and Radosavovic, I. and Ramakrishnan, S. K. and Ryan, F. and Sharma, J. and Wray, M. and Xu, M. and Xu, E. Z. and Zhao, C. and Bansal, S. and Batra, D. and Cartillier, V. and Crane, S. and Do, T. and Doulaty, M. and Erapalli, A. and Feichtenhofer, C. and Fragomeni, A. and Fu, Q. and Gebreselasie, A. and González, C. and Hillis, J. and Huang, X. and Huang, Y. and Jia, W. and Khoo, W. and Kolář, J. and Kottur, S. and Kumar, A. and Landini, F. and Li, C. and Li, Y. and Li, Z. and Mangalam, K. and Modhugu, R. and Munro, J. and Murrell, T. and Nishiyasu, T. and Price, W. and Ruiz, P. and Ramazanova, M. and Sari, L. and Somasundaram, K. and Southerland, A. and Sugano, Y. and Tao, R. and Vo, M. and Wang, Y. and Wu, X. and Yagi, T. and Zhao, Z. and Zhu, Y. and Arbeláez, P. and Crandall, D. and Damen, D. and Farinella, G. M. and Fuegen, C. and Ghanem, B. and Ithapu, V. K. and Jawahar, C. V. and Joo, H. and Kitani, K. and Li, H. and Newcombe, R. and Oliva, A. and Park, H. S. and Rehg, J. M. and Sato, Y. and Shi, J. and Shou, M. Z. and Torralba, A. and Torresani, L. and Yan, M. and Malik, J.},
  title     = {Ego4D: Around the World in 3,000 Hours of Egocentric Video},
  booktitle = {Proceedings of the IEEE/CVF Conference on Computer Vision and Pattern Recognition (CVPR)},
  year      = {2022}
}

@inproceedings{xarles2024tdeed,
  title={T-DEED: Temporal-Discriminability Enhancer Encoder-Decoder for Precise Event Spotting in Sports Videos},
  author={Xarles, Albert and Escalera, Sergio and Moeslund, Thomas B. and Clap{\'e}s, Albert},
  booktitle={Proceedings of the IEEE/CVF Conference on Computer Vision and Pattern Recognition (CVPR)},
  pages={3410--3419},
  year={2024}
}

@inproceedings{lea2017tcn,
  title={Temporal Convolutional Networks for Action Segmentation and Detection},
  author={Lea, Colin and Flynn, Michael D. and Vidal, Ren{\'e} and Reiter, Austin and Hager, Gregory D.},
  booktitle={Proceedings of the IEEE Conference on Computer Vision and Pattern Recognition (CVPR)},
  pages={156--165},
  year={2017}
}

@inproceedings{lin2014coco,
  title     = {Microsoft COCO: Common Objects in Context},
  author    = {Lin, Tsung-Yi and Maire, Michael and Belongie, Serge and Hays, James and Perona, Pietro and Ramanan, Deva and Doll{\'a}r, Piotr and Zitnick, C. Lawrence},
  booktitle = {Computer Vision -- ECCV 2014},
  editor    = {Fleet, David and Pajdla, Tomas and Schiele, Bernt and Tuytelaars, Tinne},
  pages     = {740--755},
  year      = {2014},
  publisher = {Springer},
  address   = {Zurich, Switzerland}
}

@inproceedings{sun2020view,
  title     = {View-Invariant Probabilistic Embedding for Human Pose},
  author    = {Sun, Jennifer J. and Zhao, Jiaping and Chen, Liang-Chieh and Schroff, Florian and Adam, Hartwig and Liu, Ting},
  booktitle = {Computer Vision -- ECCV 2020},
  editor    = {Vedaldi, Andrea and Bischof, Horst and Brox, Thomas and Frahm, Jan-Michael},
  pages     = {53--70},
  year      = {2020},
  publisher = {Springer},
  address   = {Glasgow, UK}
}

@inproceedings{teed2020raft,
  title     = {RAFT: Recurrent All-Pairs Field Transforms for Optical Flow},
  author    = {Teed, Zachary and Deng, Jia},
  booktitle = {Computer Vision -- ECCV 2020},
  editor    = {Vedaldi, Andrea and Bischof, Horst and Brox, Thomas and Frahm, Jan-Michael},
  pages     = {402--419},
  year      = {2020},
  publisher = {Springer},
  address   = {Glasgow, UK}
}

@inproceedings{deng2009imagenet,
  title     = {ImageNet: A Large-Scale Hierarchical Image Database},
  author    = {Deng, Jia and Dong, Wei and Socher, Richard and Li, Li-Jia and Li, Kai and Fei-Fei, Li},
  booktitle = {IEEE Conference on Computer Vision and Pattern Recognition (CVPR)},
  pages     = {248--255},
  year      = {2009},
  publisher = {IEEE},
  address   = {Miami, FL, USA}
}

\clearpage
\renewcommand{\appendixtocname}{Appendix A - Attention-Based Fusion Experiments}
\renewcommand{\appendixpagename}{Appendix A - Attention-Based Fusion Experiments}

% Page numbers in the format A-1, A-2, ...
\renewcommand{\thepage}{A-\arabic{page}}
\setcounter{page}{1}

\begin{appendices}
\chapter{Attention-Based Fusion Experiments}
\label{subsec:attention_fusion}

Introducing self-attention in the first stage of MD-FED over skeleton features results in severe overfitting, as evidenced by the rapid convergence of the training loss to zero while the validation and test Edit scores remain negligible.

In the event detector module of Stage~III in MD-FED, features extracted from the optical flow and RGB branches were initially fused via element-wise addition. To enable richer inter-modal interactions, this fusion scheme was replaced with a cross-attention mechanism. However, as shown in Table~\ref{tab:fusion-strategies}, both 2-head and 8-head cross-attention configurations significantly underperform the original additive fusion. These results suggest that, for this task, simple additive fusion remains the most effective strategy, likely due to its stability and lower tendency to overfit when combining motion and appearance cues.

\begin{table}[H]
  \centering
  \caption{Comparison of fusion strategies for Stage~III of MD-FED. Simple additive fusion outperforms cross-attention variants, likely due to stability and reduced overfitting.}
  \label{tab:fusion-strategies}
  \footnotesize
  \begin{tabularx}{0.6\columnwidth}{@{}l r@{}}
    \toprule
    \textbf{Fusion Strategy} & \textbf{Edit Score (\%)} \\
    \midrule
    \textbf{MD-FED (addition)} & \textbf{81.47} \\
    2-head cross attention & 55.85 \\
    8-head cross attention & 59.97 \\
    \bottomrule
  \end{tabularx}
\end{table}

\end{appendices}

\clearpage
\renewcommand{\appendixtocname}{Appendix B - Implementation Details}
\renewcommand{\appendixpagename}{Appendix B - Implementation Details}

\renewcommand{\thepage}{B-\arabic{page}}
\setcounter{page}{1}

\begin{appendices}
\chapter{Implementation Details}
\label{appendix:b}

\section{Pre-processing}

Similar to~\autocite{liu2026umegnet}, in the pre-processing step, we extract 2D poses and optical flow using off-the-shelf estimators. For 2D pose estimation, we use HRNet~\autocite{sun2019deep} to detect athletes' poses through top-down estimation, utilizing all 17 COCO keypoints~\autocite{lin2014coco}. The 2D joint positions are normalized by rescaling and centering, following prior work~\autocite{sun2020view}. For optical flow extraction, we apply RAFT~\autocite{teed2020raft} to compute flow between consecutive frames. The final flow is obtained by subtracting the median, clipping values to $\pm 20$ pixels, and quantizing to 8 bits.

\section{Architecture}

For feature extraction, we use RegNet-Y-200MF (TSM)~\autocite{radosavovic2020designing,lin2019tsm} for the RGB and optical flow branches, and STGCN++~\autocite{duan2022pyskl} for the skeleton branch. Although these feature extractors do not always correspond to the strongest standalone backbone choices reported in the comparison tables, our objective is not to exhaustively fine-tune the backbone design or hyperparameters for maximum single-model performance. Instead, we focus on demonstrating that the proposed distillation framework itself yields consistent performance gains, even when the underlying feature extractor choice is not fully optimal. This makes the observed improvements more attributable to the framework design rather than to aggressive backbone tuning. The encoder is initialized with pretrained ImageNet-1K weights~\autocite{deng2009imagenet} and fine-tuned to meet dataset-specific needs. For optical flow input, the first convolution layer is modified to accommodate two input channels. 

\section{Training}

For event detection, the model performs dense per-frame prediction with a coarse binary localization branch and a fine-grained event classification branch. To reduce the effect of foreground-background imbalance, the coarse localization loss uses a fivefold higher weight for the foreground class than for the background class. The network is optimized with AdamW, using an initial learning rate of 0.001, three warm-up epochs with linear scaling, and cosine annealing thereafter. In this implementation, Stage~1 is a semi-supervised skeleton pretraining stage in which only $k$ clips of the training data are treated as labeled; pseudo-label training is introduced after epoch 30, with the pseudo-label weight gradually ramped up when this phase begins, in total running for 100 epochs. Stage~2 performs multimodal distillation by aligning RGB and optical-flow features to the skeleton branch using MSE loss, running for 50 epochs, while Stage~3 performs supervised fine-tuning on the $k$-clip labeled data, running for 30 epochs. Training is conducted on an A100-40G GPU.

\end{appendices}

\end{document}